%% file: main.tex
\PassOptionsToPackage{dvipsnames}{xcolor}
\documentclass[sigconf]{aamas} 

\usepackage{balance} 

\usepackage{ragged2e}
\usepackage[most]{tcolorbox}
\setcopyright{ifaamas}
\acmConference[AAMAS '25]{Proc.\@ of the 24th International Conference
on Autonomous Agents and Multiagent Systems (AAMAS 2025)}{May 19 -- 23, 2025}
{Detroit, Michigan, USA}{A.~El~Fallah~Seghrouchni, Y.~Vorobeychik, S.~Das, A.~Nowe (eds.)}
\copyrightyear{2025}
\acmYear{2025}
\acmDOI{}
\acmPrice{}
\acmISBN{}

\acmSubmissionID{<<EasyChair Submission ID>>}

\title[LLM simulation]{LLM-based Agent Simulation for Maternal Health Interventions: Uncertainty Estimation and Decision-focused Evaluation}

\author{Sarah Martinson}
\affiliation{
  \institution{Harvard University}
  \city{}
  \country{}}
\email{sarahmartinson@g.harvard.edu}

\author{Lingkai Kong}
\affiliation{
  \institution{Harvard University}
  \city{}
  \country{}}
\email{lingkaikong@g.harvard.edu}

\author{Cheol Woo Kim}
\affiliation{
  \institution{Harvard University}
  \city{}
  \country{}}
\email{cwkim@g.harvard.edu}

\author{Aparna Taneja}
\affiliation{
  \institution{Google Research India}
  \city{}
  \country{}}
\email{aparnataneja@google.com}

\author{Milind Tambe}
\affiliation{
  \institution{Harvard University}
  \city{}
  \country{}}
\email{milind_tambe@harvard.edu}

\begin{abstract}

Agent-based simulation is crucial for modeling complex human behavior, yet traditional approaches require extensive domain knowledge and large datasets. In data-scarce healthcare settings where historic and counterfactual data are limited, large language models (LLMs) offer a promising alternative by leveraging broad world knowledge. This study examines an LLM-driven simulation of a maternal mobile health program, predicting beneficiaries’ listening behavior when they receive health information via automated messages (control) or live representatives (intervention). Since uncertainty quantification is critical for decision-making in health interventions, we propose an LLM epistemic uncertainty estimation method based on binary entropy across multiple samples. We enhance model robustness through ensemble approaches, improving F1 score and model calibration compared to individual models. Beyond direct evaluation, we take a decision-focused approach, demonstrating how LLM predictions inform intervention feasibility and trial implementation in data-limited settings. The proposed method extends to public health, disaster response, and other domains requiring rapid intervention assessment under severe data constraints. All code and prompts used for this work can be found at \url{https://github.com/sarahmart/LLM-ABS-ARMMAN-prediction}.

\end{abstract}


\newcommand{\BibTeX}{\rm B\kern-.05em{\sc i\kern-.025em b}\kern-.08em\TeX}

\begin{document}


\pagestyle{fancy}
\fancyhead{}


\maketitle 




\input{Sections/1-intro}

\input{Sections/2-related_works}

\input{Sections/3-method}

\input{Sections/4-results}

\input{Sections/5-conclusion}






\bibliographystyle{ACM-Reference-Format} 
\bibliography{main}


\end{document}

%% file: Sections/1-intro.tex
\section{Introduction}

Developing and deploying effective healthcare interventions in under-served regions often requires substantial time and resource investments. However, the absence of historical data or prior evaluations often makes it difficult to assess the efficacy of new health programs. Health workers and program managers may identify promising interventions, but the high costs of running trials or large-scale implementations complicate prioritization. To address this, we explore the potential of large language models (LLMs) for early-stage intervention assessment in data-deficient contexts, where behavioral data is unavailable and decision-making relies solely on contextual, population, and demographic information.

LLMs encode vast amounts of world knowledge, allowing them to approximate counterfactual agent-based predictions of individual behavior under hypothetical interventions. While these predictions do not constitute causal counterfactuals, they can provide plausible behavioral simulations based on encoded sociodemographic priors \cite{yang2024large}. Our framework leverages these capabilities to assess maternal health programs before implementation.

Our study focuses on ARMMAN, a maternal health non-governmental organization in India that operates the mMitra program \cite{mmitra}, an automated weekly call service delivering health messages to pregnant women and new mothers. While the programme has demonstrated success in improving health outcomes, sustaining engagement remains challenging \cite{mmitra}. Live service calls by health workers are used to support high-risk mothers, but resource constraints necessitate precise targeting. Currently, this targeting process follows a two-stage predictive framework: first, training a model to predict listening behavior, then using a Restless Multi-Armed Bandit \cite{verma2023restless} to allocate resources. While this approach effectively assigns live service calls, it requires extensive historical training data. There is no solution for low-data settings, such as evaluating interventions in new locations or under novel conditions. To explore whether LLM-powered agent-based predictions can provide robust appraisals in such settings, this study reimagines the existing mMitra program as an intervention \textit{not yet implemented}. Given contextual details about the program and sociodemographic characteristics of potential participants, we prompt an LLM acting as a mother (agent) to predict binary engagement.

Different LLMs possess varying world knowledge, make predictions with different levels of uncertainty, and encode distinct biases. To enhance robustness and model calibration, we compare multiple LLMs and ensemble their predictions. Inspired by team-based agent collaboration frameworks \cite{marcolino2013teamformation, marcolino2014hardproblem}, we evaluate three ensembling strategies: direct averaging, epistemic uncertainty-weighted aggregation, and lowest-uncertainty prediction selection. We investigate their ability to balance robustness and calibration while mitigating biases in individual models, and assess their effectiveness in terms of predictive accuracy, F1 score, and calibration. Additionally, we adopt a decision-focused approach to evaluate how well LLM-based predictions support real-world decisions about whether to implement interventions in data-deficient contexts.


\subsection*{Contributions}

We (1) introduce an LLM-based approach for prioritizing interventions and estimating their impact in the absence of historical behavioral data, and when real-world experimentation is limited or infeasible; (2) comprehensively compare and evaluate multiple ensembling methods for LLM prediction, demonstrating their effects on accuracy, F1 score, and log-likelihood; and (3) introduce a decision-focused evaluation pipeline leveraging LLM predictions and counterfactual modeling to guide intervention assessment in resource-constrained social good settings. While our primary focus is maternal health, the framework is generalizable to other domains requiring rapid, data-efficient decision-making, such as disaster response, pandemic control, and targeted social programs. 


%% file: Sections/2-related_works.tex
\section{Related Work}

\paragraph{Agent-Based Modeling for Social Good}
Agent-based modeling (ABM) has been widely used for public health, social welfare, and resource allocation \cite{zhao2024towards, jain2024irl}. Successful applications of automated agents include pilot simulations for battlefield environments \cite{tambe1995intelligent}, airport evacuation modeling \cite{tsai2011escapes}, training systems for disaster incident commanders \cite{schurr2006using}, and task allocation and discovery for dynamic disaster response \cite{nair2002task}. ABM has also demonstrated success in real implementations of maternal mobile health (mHealth) programs \cite{verma2024leveraging}. These models allow researchers and policymakers to experiment with virtual populations, identifying how different interventions might affect outcomes \cite{jain2024irl}. However, traditional ABMs typically rely on hand-crafted rules and historical time-series data to model agent behavior. In data-deficient settings, this dependence on dense behavioral records poses a significant challenge, as real-world datasets are often sparse or incomplete.

Recent advances in foundation models and LLMs suggest a new paradigm of simulation agents that encode extensive world knowledge \cite{zhao2024towards}. Unlike handcrafted ABMs, LLM-based agents can generate counterfactual behavioral predictions even in low-data environments, offering more flexible and adaptive simulations. This capability is pertinent in AI for social impact contexts, where traditional modeling pipelines are often labor- and resource-intensive and domain-specific, limiting applicability and scalability \cite{zhao2024towards}. Additionally, \cite{gao2024large} discusses the integration of LLMs into ABM across various fields, including public health and social welfare. By incorporating LLMs, ABMs can simulate complex human behavior and interactions more effectively, even in data-sparse environments. However, limitations of LLM-based simulations, including ensuring their reliability and interpretability, and their real-world effectiveness remain underexplored \cite{gao2024large}. Future research must validate these models through empirical studies to assess their reliability and robustness in real decision-making scenarios.

Predicting engagement in maternal healthcare initiatives relevant for large-scale mHealth programs \cite{verma2024leveraging}. Although models like Markovian restless bandits \cite{verma2023restless, verma2024leveraging} can optimize intervention resources, these approaches face limitations in non-Markovian contexts, particularly with multiple interventions or varied user behavior \cite{verma2024leveraging}. Moreover, standard ABMs offer little insight into counterfactual outcomes, as they rely on observed behaviour histories. In contrast, LLM agent simulations can hypothesize novel interactions by encoding prior world knowledge. 


\paragraph{LLMs for Human Behavior Prediction}
LLMs have demonstrated considerable potential for approximating human behavior by capturing language comprehension, cognitive heuristics, and common human systematic biases \cite{huijzer2023large}. Recent studies have explored LLMs as proxies in social experiments, using value injection fine-tuning techniques to predict opinions \cite{kang-etal-2023-values}, constructing simulacra that can produce personified responses for given characters \cite{xie2024human}, and demonstrating that LLMs perform better than traditional cognitive models in predicting human behavior in sequential decision-making tasks \cite{nguyen2024predicting}. These advances hint at the potential for using LLMs to predict behavior of participants in health-related trials. However, challenges persist in calibrating LLM outputs, addressing overconfidence, and managing bias and variance \cite{gao2024large}. 

\paragraph{LLM prediction ensembling}
Previous work highlights advantages of including diverse agents in multi-agent teams, demonstrating that a collection of individually weaker agents can outperform uniform teams of individually superior agents \cite{marcolino2013teamformation, marcolino2014hardproblem, jiang2014diverse}. Recent studies extend this to LLM ensembles, showing how techniques such as maximizing diversity \cite{tekin2024llmtopla}, using sampling-and-voting methods at scale \cite{li2024more}, and pairwise ranking with generative fusion \cite{jiang-etal-2023-llm} can yield notable performance gains over component LLM models. Building on these insights, we harness ensembling to stabilize and refine LLM-based behavior predictions, targeting applications in maternal health programs where reliable and diverse agent perspectives are important in guiding effective interventions. 

\paragraph{Uncertainty Estimation of LLM predictions.}
Uncertainty quantification is vital in high-stakes settings where inaccurate predictions may lead to ineffective or even harmful outcomes. While accurate confidence estimates enable more reliable decision-making—where a model's certainty should be correlated with its correctness \cite{xiong2024llmuncertainty}—current LLM prediction methods often overlook uncertainty or fail to incorporate it systematically in ensembling \cite{kong2020calibrated}. Recent efforts distinguish epistemic uncertainty—reflecting gaps in model knowledge—from aleatoric uncertainty, arising from entropy in the underlying data distribution \cite{ahdritz2024, abbasi-yadkori2024to}. These studies propose iterative prompting techniques to approximate uncertainties from LLM predictions. Since LLMs predict tokens from a vast textual corpus, they must manage both inherent randomness in language and data (aleatoric uncertainty), and gaps in their own knowledge (epistemic uncertainty). This suggests LLMs naturally contain internal representations of uncertainty that can be estimated to provide indications of model confidence \cite{ahdritz2024}. 


%% file: Sections/3-method.tex
\section{Method}

\subsection{Data, Models \& Prompting}

\paragraph{Data} We have access to anonymized data from ARMMAN's mMitra program on 3000 mothers over 40 weeks. We consider two groups—a control group where all mothers receive only automated calls with health messages each week of the program, and an intervention group, where a random subset of mothers receives a live call from a health worker instead of the automated message for a specific week. Intervention calls convey the same information as automated messages, tailored to the mother’s stage of pregnancy or postpartum period. No mother receives an intervention more than once, and all intervention calls occur in the first six weeks.

For each mother, we consider binary actions—whether a mother received a live call or not—and corresponding continuous states—weekly listening times to health messages. To define engagement, listening times are converted to binary engagement states, where engagement corresponds to listening to a message for more than 30 seconds, while listening less (including non-answering) is considered unengaged for that week\footnote{This aligns with engagement criteria used in previous evaluations of mMitra \cite{verma2023restless, verma2024leveraging}.}. The LLMs' prediction task is to classify whether a mother will engage or not in a given week.

Each mother is associated with a set of numerically encoded sociodemographic features (Table~\ref{tab:sociodemographics}). We assume contextual information about the program and a general population of mothers who may enroll is available. Specifically, we assume mothers are identified through maternal health clinic visits and provide participation consent. In our framework, LLMs act as mothers, making engagement predictions based on given sociodemographic profiles and program details, leveraging prior world knowledge about health interventions, telehealth adoption, and behavioral responses to generate predictions regarding a mother's engagement.

\begin{table*}[ht]
    \footnotesize
    \centering
    \begin{tabular}{l|p{10cm}}
        \toprule
        \textbf{Category} & \textbf{Available Characteristics} \\
        \midrule
        Enrollment Information & gestational age at enrollment, delivery status \\
        Reproductive History & gravidity, parity, number of stillbirths, number of living children \\
        Age Groups & \texttt{<20}, 20--24, 25--29, 30--34, 35+ \\
        Language & Hindi, Marathi, Kannada, Gujarati, English \\
        Education Level & Illiterate, 1--5 years, 6--9 years, 10th pass, 12th pass, graduate, postgraduate \\
        Phone Ownership & mother’s phone, husband’s phone, family phone \\
        Preferred Call Time & 8:30--10:30, 10:30--12:30, 12:30--15:30, 15:30--17:30, 17:30--19:30, 19:30--21:30 \\
        Enrollment Channel & community enrollment, hospital enrollment, ARMMAN enrollment \\
        Monthly Income (INR) & 0--5000, 5001--10000, 10001--15000, 15001--20000, 20001--25000, 25001--30000 \\ 
        \bottomrule
    \end{tabular}
    \caption{Sociodemographic characteristics available for mothers in the program.}
    \label{tab:sociodemographics}
\end{table*}

\paragraph{Models and hyper-parameters}
For this study, we evaluate a selection of heavyweight and lightweight LLMs from Google, OpenAI, and Anthropic to capture a representative range of closed-source LLM capabilities. Heavyweight models include Gemini 1.5 Pro \cite{google2024gemini, geminipro} and GPT-4o \cite{gpt4o}, while lightweight models include Gemini 1.5 Flash \cite{geminiflash}, GPT-4o mini \cite{gpt4omini}, and Claude Instant 1.2 \cite{claudeinstant}. Hyperparameters for each model are detailed in Table~\ref{tab:hyperparams}.

\begin{table*}[ht]
    \footnotesize
    \centering
    \begin{tabular}{l|l|l|l}
        \toprule
        \textbf{Provider} & \textbf{Weight} & \textbf{Model} & \textbf{Generation Setup} \\
        \midrule
        Google & Heavy & Gemini 1.5 Pro & \texttt{model = gemini-1.5-pro-002, temperature = 1, max\_tokens = 8192} \\
        Google & Light & Gemini 1.5 Flash & \texttt{model = gemini-1.5-flash-002, temperature = 1, max\_tokens = 8192} \\
        OpenAI & Heavy & GPT-4o & \texttt{model = gpt-4o, temperature = 0.7, max\_tokens = 2048} \\
        OpenAI & Light & GPT-4o mini & \texttt{model = gpt-4o-mini, temperature = 0.7, max\_tokens = 2048} \\
        Anthropic & Light & Claude Instant 1.2 & \texttt{model = claude-instant-v1, temperature = 0.7, max\_tokens = 2048} \\
        \bottomrule
    \end{tabular}
    \caption{Generating hyperparameters for various LLMs.}
    \label{tab:hyperparams}
\end{table*}


\paragraph{Prompting}
We evaluate multiple intervention and control scenarios, described in Section~\ref{sec:sim_scenrios}. In each scenario, all models receive the same set of five prompts per mother per time step. Prompts vary slightly in wording but convey the same core information, including a description of the mHealth program, the sociodemographic characteristics of the mother, and a request for a prediction regarding her engagement at that time step. 

All prompts state that simulation is weekly and specify the mode of message delivery—as a brief automated message or live service call from a health worker containing the same information. A sample prompt is provided in Box 1\footnote{For the full set of prompts used in both intervention and control scenarios, see \href{https://github.com/sarahmart/LLM-ABS-ARMMAN-prediction/blob/main/models/prompt_templates.py}{here}.}. Characteristics are listed in Table~\ref{tab:sociodemographics}, and key differences between control (orange) and intervention (green) prompts are highlighted. Since minor variations in prompt wording can affect LLM outputs \cite{huijzer2023large, sclar2024quantifying}, we keep differences between intervention and control versions of each prompt minimal to maintain comparability. Each model is queried five times per prompt, yielding 25 predictions per mother per time step.

\subsection{Simulation Scenarios} \label{sec:sim_scenrios}

We examine three simulation settings: (1) \textit{intervention}, (2) \textit{counterfactual}, and (3) \textit{control}. These allow us to evaluate intervention impact, compare predicted outcomes under counterfactual conditions, and establish a baseline for engagement without intervention.


\begin{tcolorbox}[breakable, enhanced, colframe=gray!60!purple, colback=gray!5!white, title=\textbf{Box 1:} Sample LLM Prompt \\ \footnotesize{\textcolor{orange}{\textbf{<no intervention version>}}\textcolor{LimeGreen}{\textbf{<intervention version>}}}]

\small{You are a mother enrolled in the ARMMAN Maternal and Child Healthcare Mobile Health program. 
ARMMAN is a non-governmental organization in India dedicated to reducing maternal and neonatal mortality among underprivileged communities. 
Through this program, you receive weekly preventive health information via brief 
\textcolor{orange}{\textbf{<automated voice messages>}} \textcolor{LimeGreen}{\textbf{<phone calls>}}. 

In this simulation, each time step represents one week.

Below is your background and history with the program. 

\textbf{Your Background:}
\begin{itemize}
    \itemindent=-15pt
    \raggedright
    \item You enrolled in the program during the \texttt{\{enroll\_gest\_age\}} week of your pregnancy.
    \item You are \texttt{\{age\_category\}} years old.
    \item Your family's monthly income is INR \texttt{\{income\_bracket\}}.
    \item Your education level is \texttt{\{education\_level\}}.
    \item You speak \texttt{\{language\}}.
    \item You own a \texttt{\{phone\_owner\}} phone.
    \item You prefer receiving calls during \texttt{\{call\_slot\_preference\}}.
    \item You enrolled in the program through \texttt{\{channel\_type\}}.
    \item You are currently in the \texttt{\{enroll\_delivery\_status\}} stage.
    \item You have been pregnant \texttt{\{g\}} times, with \texttt{\{p\}} successful births.
    \item You have had \texttt{\{s\}} stillbirth(s) and have \texttt{\{l\}} living child(ren).
\end{itemize}

\textbf{Past Behavior:} The following is a record of your previous listening behavior (each representing one week):
\texttt{\{past\_behavior\}}

Based on this information, as well as the context of the program and on typical behavior of mothers in India, decide whether you will be engaged with the next automated health message.

\textbf{Key Consideration:} Engagement at a previous week does not necessarily imply engagement at the next, and lack of engagement at a previous week does not necessarily imply future lack of engagement. 
Engagement should depend on your specific circumstances that week (e.g. need for reassurance or information, phone availability, schedule, etc.).
Being unable to answer a call that week implies a lack of engagement for that week.

\textbf{Question: }Will you be engaged with the next \textbf{\textcolor{orange}{<automated health message>}\textcolor{LimeGreen}{<call from a health worker>}}?

Please respond with your final decision in the format: `\#\#Yes\#\#' for engagement or `\#\#No\#\#' for lack of engagement in this week. 
Your response need only contain one of the following: `\#\#Yes\#\#' OR `\#\#No\#\#'. No other text should be included. 
}
\end{tcolorbox}

\paragraph{Intervention} 

We select a sample with $\sim 60\%$ of mothers receiving an intervention at some point during the 40-week period, reflecting real program constraints. This corresponds to $\sim 10\%$ of the total population receiving a live call in each of the first six program weeks. This weekly threshold reflects a feasible allocation in resource-constrained programs, with an average of 0.015 live calls per mother over the entire population and program duration. LLMs simulating mothers who receive the intervention are prompted with the \textbf{intervention} prompt (Box 1) for the corresponding week. LLMs simulating mothers who do not receive the intervention are provided with the \textbf{no intervention} prompt, which specifies an automated telehealth message instead of a live call. At each time step, the LLM generates a binary engagement prediction.

We analyze two subsamples from this group: (1) a larger sample of 500 mothers, used to evaluate predictive performance of LLMs in an intervention context, and (2) a smaller representative subsample of 100 mothers, selected using K-means clustering to match the larger sample in terms of normalized feature values, mean engagement, and linear engagement trends. The smaller subsample is used to compare intervention effects across the three settings.

\paragraph{Counterfactual} 
We use the same representative subsample of 100 mothers as in the \textit{intervention} group. In this scenario, all mothers receive \textbf{no intervention} prompts containing engagement history but omitting information regarding past interventions, allowing us to assess predicted engagement in the absence of explicit actions.

\paragraph{Control}
This group consists of 100 distinct mothers who never received an intervention. All LLMs simulating these mothers receive \textbf{no intervention} prompts. However, in this case, action history is included and consists entirely of zero-action trajectories, reinforcing the absence of live calls.

\subsection{Prediction Ensembling}

For each simulation setting, we run all models in Table~\ref{tab:hyperparams}, generating $N = 25$ predictions per mother per time step. We compute mean predictions and associated epistemic uncertainty at each time step.

\paragraph{Epistemic Uncertainty}

Epistemic uncertainty~\cite{kong2020sde, limuben} captures uncertainty in the model’s knowledge and is distinct from aleatoric uncertainty, which arises from inherent randomness in data. Properly distinguishing between these sources of uncertainty allows us to weight predictions based on model confidence. Using $N$ repeated binary predictions and methods from \cite{kong2023uq}, we quantify predictive uncertainty for a given mother and time step as the binary entropy $H$ of mean predictions across queries. 

Let $p_i$ be the individual predictions from a single model, and $\bar p = \frac 1 N \sum_{i=1}^N p_i$ be their mean. Predictive uncertainty is then
$$H(\bar p) = -\bar p \log(\bar p) - (1 - \bar p) \log(1-\bar p).$$
Aleatoric uncertainty is the mean of individual prediction entropies
$$\frac 1 N \sum_{i=1}^N H(p_i) = \frac 1 N \sum_{i=1}^N [-p_i \log(p_i) - (1-p_i)\log(1-p_i)],$$
and epistemic uncertainty is then obtained as the difference
$$H(\bar p) - \frac 1 N \sum_{i=1}^N H(p_i).$$
This provides a per-mother, per-time-step uncertainty estimate. 

\paragraph{Ensembling} \label{sec:ensembling}

We evaluate three ensembling strategies:

\paragraph{(1) Direct averaging} 
We compute mean predictions across models, presenting this average as the ensemble’s final prediction probability (and binarizing where relevant). All models contribute equally to the ensemble, regardless of uncertainty. 


\paragraph{(2) Uncertainty-weighted aggregation}
We employ a Bayesian weighting mechanism using epistemic uncertainty to determine each model’s contribution to the aggregated prediction. Lower epistemic uncertainties $u_j$ for model $j$ correspond to higher model precisions $\tau_j = \frac 1 {u_j}$ and greater influence in the aggregated result. For each time step, the combined prediction is a weighted average 
$$ p_{\text{combined}} = \frac{\sum_{j=1}^M \tau_j \bar p_j}{\sum_{j=1}^M \tau_j},$$
where $\bar p_j$ is the mean prediction from model $j$, $\tau_j$ is the corresponding precision, and $M$ is the number of models in the ensemble. This ensures models with higher confidence exert greater influence on the final prediction.


Since different models may estimate epistemic uncertainty on different scales, we apply rank normalization across models before aggregation. For each model  $j$  with uncertainty estimate  $u_j$ , we compute the normalized uncertainty
$$u_j^\prime = \frac{\text{Rank}(u_j)}{\max(\text{Rank}(u_1, u_2, \dots, u_M))},$$
where $\text{Rank}(u_j)$  assigns a rank to each uncertainty relative to uncertainties of all $M$ models. This preserves relative ordering while ensuring values are normalized to the same scale, preventing any single model’s uncertainty from dominating the ensemble.

\paragraph{(3) Lowest-uncertainty selection}
As a baseline, we use rank-normalized epistemic uncertainty values to select, for each mother at each time step, the model's prediction corresponding with the lowest uncertainty $p_{\text{selected}} = p_{\arg \min u_j}.$

\subsection{Evaluation}

\paragraph{Direct evaluation}

For the larger group of 500 mothers, we simulate engagement trajectories over the full 40-week period for each LLM. This larger sample allows for a comprehensive assessment of model performance across diverse engagement trajectories. We compute three ensemble prediction values and conduct direct evaluations of component and ensemble predictions over time using accuracy, F1 score, and log-likelihood. Additionally, we perform a bias analysis for all component models and ensembles by breaking accuracy down by sociodemographic group (participant age, income, education level, and language). 

\paragraph{Decision-focused Analysis}

Once individual model and ensemble performance have been established, we prompt the LLMs for 15 weeks on the 100-mother subsamples of the \textit{intervention}, \textit{counterfactual}, and \textit{control} groups. For each setting, we compute predicted engagement trajectories over time to determine whether the intervention setting exhibits increased engagement relative to no-intervention settings. Additionally, we analyze prediction transition probabilities over time to assess how LLMs model changes in retention and new/re-engagement following interventions.

%% file: Sections/4-results.tex
\section{Experimental Results \& Analysis}

Throughout this section, we compare performance of individual LLMs (Google's Gemini 1.5 Flash and 1.5 Pro, OpenAI's GPT-4o and GPT-4o mini, and Anthropic's Claude Instant) to the three ensemble methods described in Section~\ref{sec:ensembling}: direct averaging (black), epistemic uncertainty-weighted aggregation (red), and lowest-uncertainty selection (gray). Ensembles are plotted for all five models, but curves are separated by provider for readability.

\subsection{Evaluation Metrics}

Here, we evaluate predictions of the LLMs for 500 mothers in the \textit{intervention} setting over a 40-week period.

\paragraph{Accuracy}

\begin{figure}[h]
    \centering
    \includegraphics[width=\linewidth]{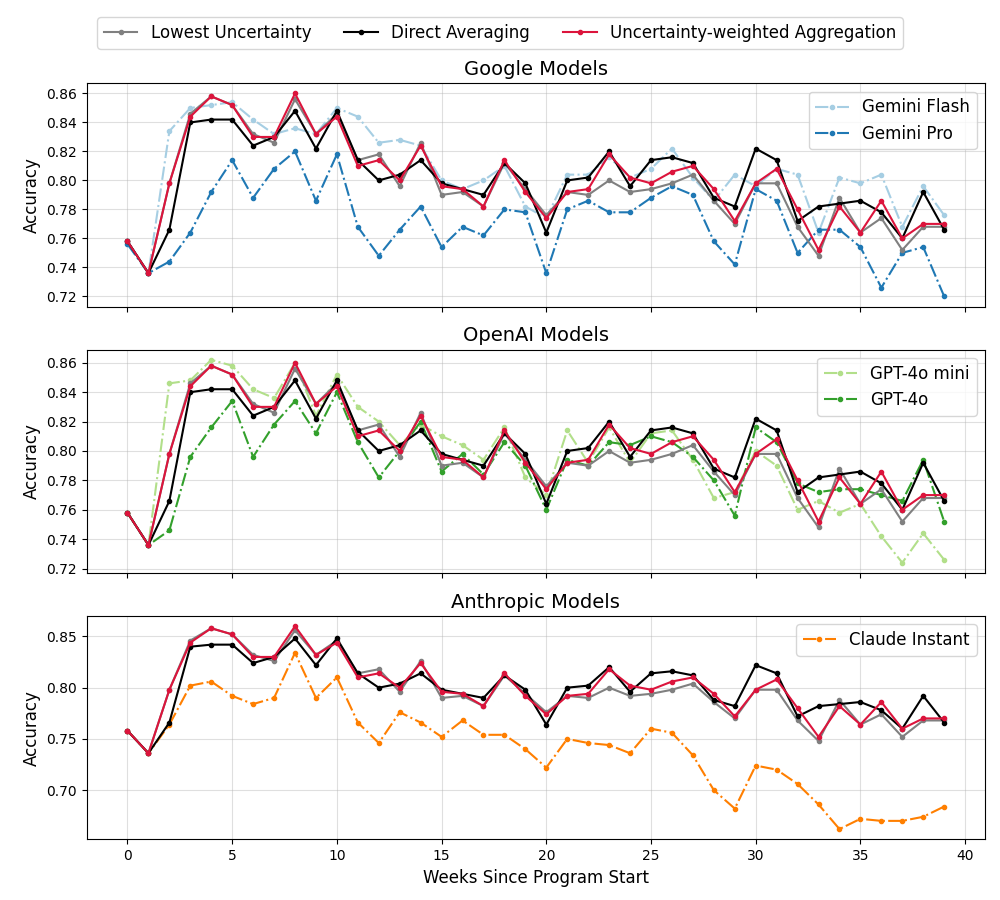}
    \caption{Mean accuracy of component and ensemble models over time, grouped by provider.}
    \Description{This figure shows the average accuracy of component and ensemble models over time, highlighting performance trends and differences in accuracy between models.}
\label{fig:acc_vs_week}
\end{figure}

In Figure~\ref{fig:acc_vs_week}, we assess predictive accuracy of individual models and ensembling methods over time, with models grouped by provider. All models exhibit a decline in accuracy with time, which is expected in an autoregressive prediction setting because of error propagation from earlier predictions, making later weeks increasingly difficult to predict. 

Claude Instant (bottom) consistently underperforms relative to other models, with accuracy exceeding 0.8 in only four weeks. By contrast, Gemini Flash and GPT-4o emerge as the strongest individual models in terms of accuracy. Gemini Flash consistently outperforms Gemini Pro, despite the latter being a more powerful model in general-purpose settings \cite{geminipro}. 
GPT-4o mini achieves strong early performance ($\sim 0.85$ accuracy) but experiences greater fluctuation and a decline after $\sim 30$ weeks.

Notably, uncertainty-weighted aggregation (red) and direct averaging (black) mitigate this decline, indicating that ensemble methods help stabilize predictions when component model performance deteriorates. Meanwhile, GPT-4o exhibits a different trend, with initially lower accuracy that improves over time, surpassing GPT-4o mini after $\sim 30$ weeks. Ensemble methods effectively track these model performance shifts, dynamically adjusting to follow the best-performing model at each stage. Overall, ensemble methods provide robustness across models, improving predictions relative to weaker models such as Claude Instant and Gemini Pro, without sacrificing accuracy from stronger models. 



\subsubsection{F1 score}

\begin{figure}[h]
    \centering
    \includegraphics[width=\linewidth]{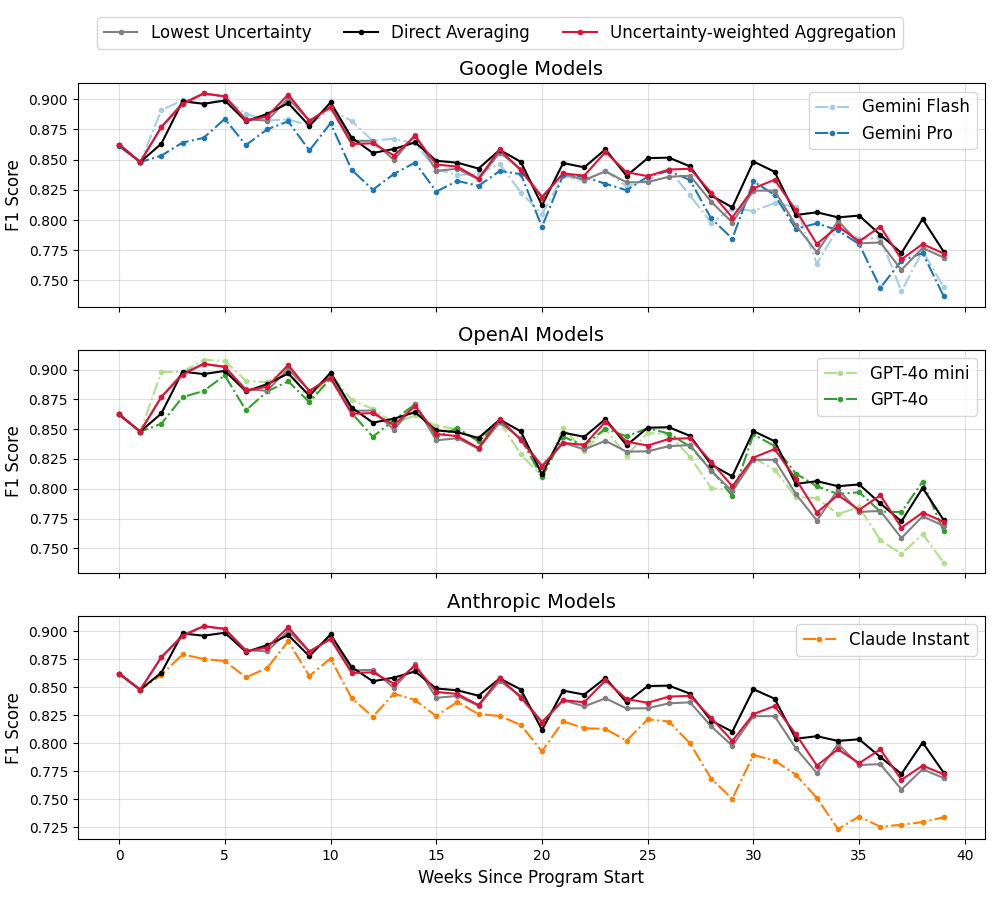}
    \caption{Mean F1 score of component and ensemble models over time, grouped by provider.}
    \label{fig:f1_vs_week}
    \Description{F1 scores follow similar trends to accuracies, starting high and declining with time.}
\end{figure}

We plot F1 score (Figure~\ref{fig:f1_vs_week}) for all models over time for a more balanced evaluation to ensure models are not simply optimizing for the majority class. In this group, the total engagement proportion is 0.59, suggesting a possible class imbalance. 

F1 scores demonstrate similar trends to accuracies. Unlike in accuracy results, Gemini Pro achieves slightly higher F1 scores than Gemini Flash. This suggests that Gemini Pro may have better recall, identifying more engagement cases, even at the cost of slightly reduced precision. GPT-4o and GPT-4o mini exhibit similar F1 score trends, while Claude Instant continues to demonstrate significantly lower performance compared to other models.

Most notably, no individual component model consistently outperforms either of the aggregation methods, reinforcing their effectiveness. By integrating predictions from different models, ensembling likely balances the precision-recall trade-off more effectively than any single model. Across all groups, uncertainty-weighted aggregation (red) and direct averaging (black) perform strongly, generally outperforming all individual models. In contrast, lowest-uncertainty selection (gray) underperforms slightly in the long term, reinforcing that selecting the single most confident prediction does not necessarily yield the best balance between precision and recall. The lower F1 scores for lowest-uncertainty selection suggest that high-confidence predictions might be biased toward precision, leading to reduced recall and classifier effectiveness.

\subsubsection{Log-likelihood}

\begin{figure}[h]
    \centering
    \includegraphics[width=\linewidth]{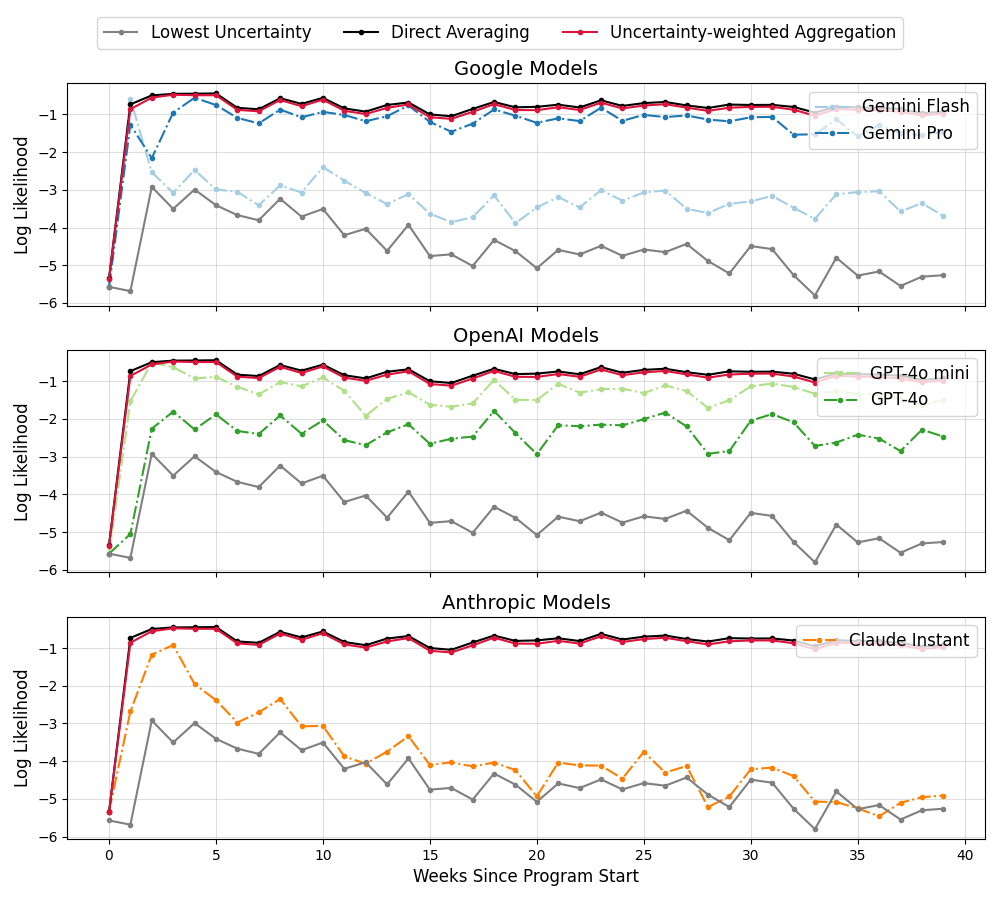}
    \caption{Mean log-likelihood of component and ensemble models over time, grouped by provider.}
    \label{fig:log-lik_vs_week}
    \Description{Log-likelihood of aggregated models is significantly better than component models, which themselves are better than lowest-uncertainty selection.}
\end{figure}

Figure~\ref{fig:log-lik_vs_week} plots model log-likelihood over time, providing insight into model confidence and calibration. Unlike accuracy and F1 score, which assess binary correctness, log-likelihood captures both correctness and model confidence in predictions. Higher log-likelihoods indicate correct classifications \textit{and} well-calibrated probability estimates, making it important for evaluating model reliability in uncertainty-aware settings such as ours.

As with other metrics, log-likelihood exhibits an initial increase across all models during early calibration, followed by stabilization and a slight decline over time. Aggregation methods demonstrate highest log-likelihood values, indicating better alignment with true probabilities than component models and suggesting that aggregation mitigates overconfidence, leading to better-calibrated probability estimates and improved reliability.

Among individual models, Gemini Pro achieves significantly higher log-likelihood than Gemini Flash, despite having lower accuracy. A similar trend is observed with GPT-4o mini, which outperforms GPT-4o in log-likelihood despite lower raw accuracy. This suggests Gemini Pro and GPT-4o mini are better calibrated than their more accurate counterparts. While Gemini Flash and GPT-4o may achieve higher accuracy by making more confident predictions, their probability estimates may be less well-calibrated, leading to lower log-likelihood scores.

Among ensemble approaches, direct averaging (black) slightly outperforms uncertainty-weighted aggregation (red) in log-likelihood, both stabilizing around -0.75 in the long term. This suggests that weighting predictions by uncertainty may amplify miscalibrated models—an overconfident but systematically biased model may contribute higher precision weightings and lead to lower overall log-likelihood. Lowest-uncertainty selection (gray) performs particularly badly, indicating poor model calibration and an unreliable ensembling strategy—this method likely systematically selects overconfident predictions, even when incorrect. 

\subsubsection{Bias Analysis} 

\begin{figure}[ht]
    \centering
    \includegraphics[width=\linewidth]{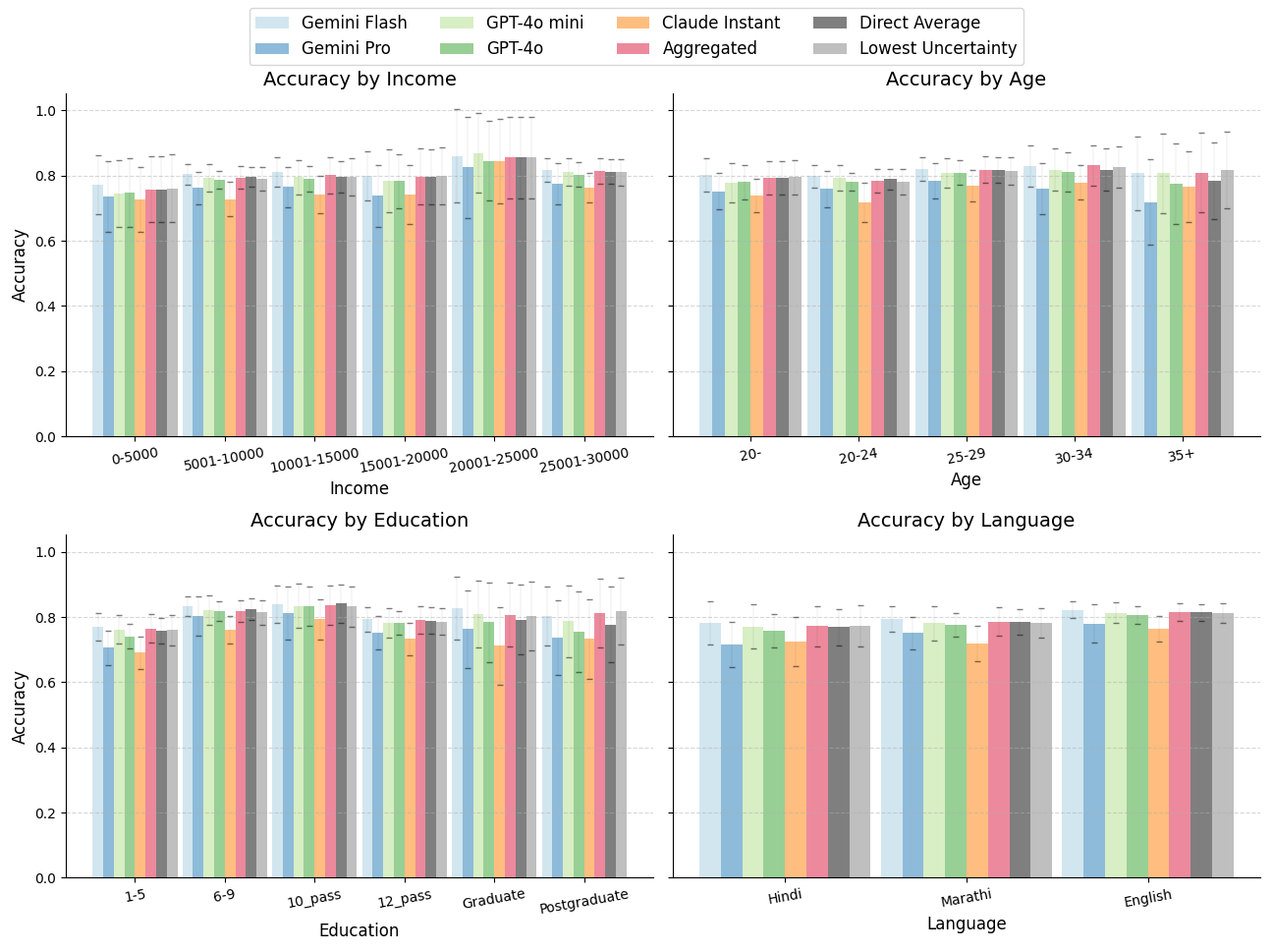}
    \caption{Average total model accuracy by sociodemographic feature for component and ensemble models.}
\label{fig:bias_acc}
\Description{Bars are all similar in height, indicating similar accuracy across models and little bias.}
\end{figure}

We assess model fairness by evaluating accuracy across sociodemographic groups. Figure~\ref{fig:bias_acc} plots accuracy for all component models and ensembles across income, age, education, and language groups. Bias is quantified as the maximum observed accuracy difference across feature categories for each model.

Accuracy remains consistent across feature groups, demonstrating minimal model bias. There is a slight trend toward higher accuracy for higher income brackets, 10-pass and 12-pass education categories, and English-speaking mothers, but no group exhibits significantly stronger performance than others, implying no strong systematic bias. Across sociodemographic groups, individual models exhibit similar accuracy distributions, with slight variations. Ensemble methods generally outperform individual models or at least match best performances, providing robust predictions across feature categories. The improved robustness of ensemble methods suggests aggregating predictions across models helps mitigate individual model biases and improve generalization.


\subsection{Decision-focused Analysis}

In this section, we focus on the two aggregation methods as they have demonstrated superior performance in accuracy, F1 score, and log-likelihood. Because of the poor performance of the lowest-uncertainty selection method, we exclude it from further analysis.

We analyze engagement predictions across the three settings using a cohort of 100 mothers over a 15-week period. We restrict analysis to this shorter time horizon, as engagement predictions become increasingly unstable beyond this point and accuracy tends to decline over time. 

\paragraph{Total Engagement}

\begin{figure}[h]
    \centering
    \includegraphics[width=\linewidth]{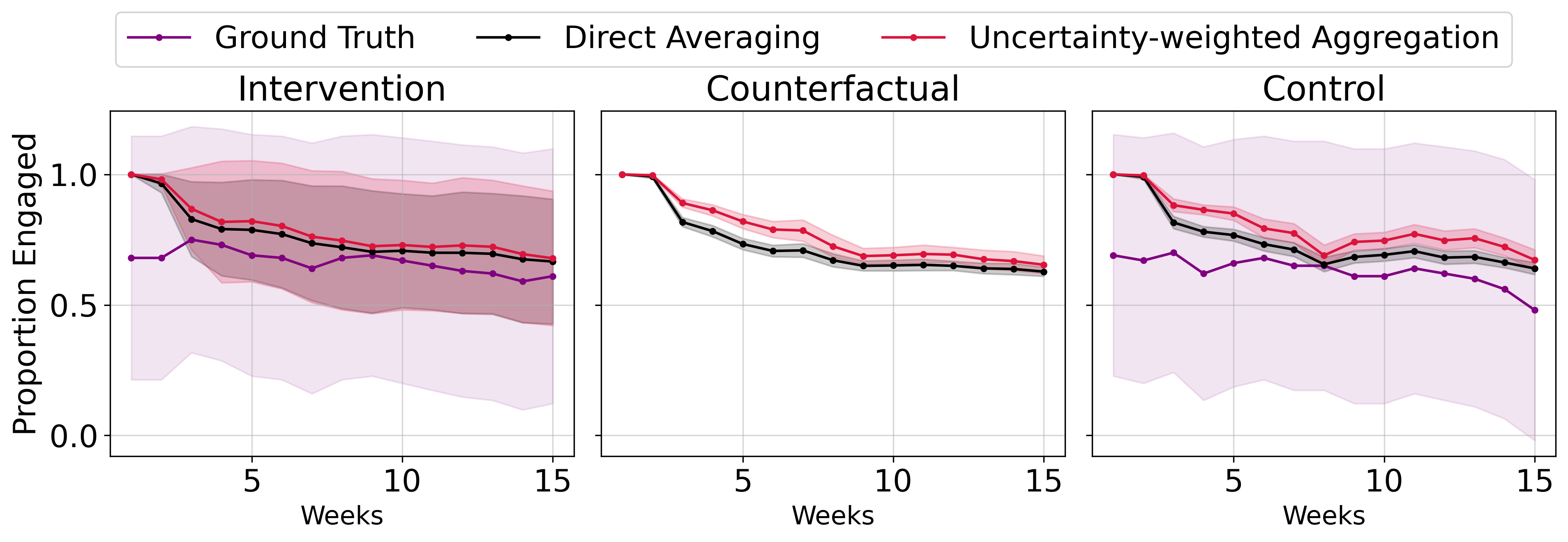}
    \caption{Mean engagement proportion over time in (left to right) the \textit{intervention}, \textit{counterfactual} and \textit{control} settings.}
    \label{fig:engagement_over_time}
    \Description{}
\end{figure}

Figure~\ref{fig:engagement_over_time} presents the mean engagement proportion over time for each setting. Note the \textit{counterfactual} setting (middle) does not include a ground truth curve, as no direct observations exist for this scenario.

Across settings, both aggregation methods (as well as individual component models, omitted for clarity) initially overpredict engagement with relatively low variance. Engagement declines over time in all cases, with uncertainty-weighted aggregation (red) consistently predicting slightly higher engagement than direct averaging (black). Both methods overestimate engagement relative to ground truth in \textit{intervention} and \textit{control} settings, particularly in the latter, suggesting systematic optimism in predictions. Despite initial bias, predictions gradually begin to align better with the observed engagement trend. In the \textit{intervention} setting, this alignment occurs around the third week, while in the \textit{control} setting, predictions this is closer to the sixth. Engagement predictions remain slightly elevated throughout the study period, indicating persistent overestimation of the ensemble methods.

\begin{figure}[h]
    \centering
    \includegraphics[width=\linewidth]{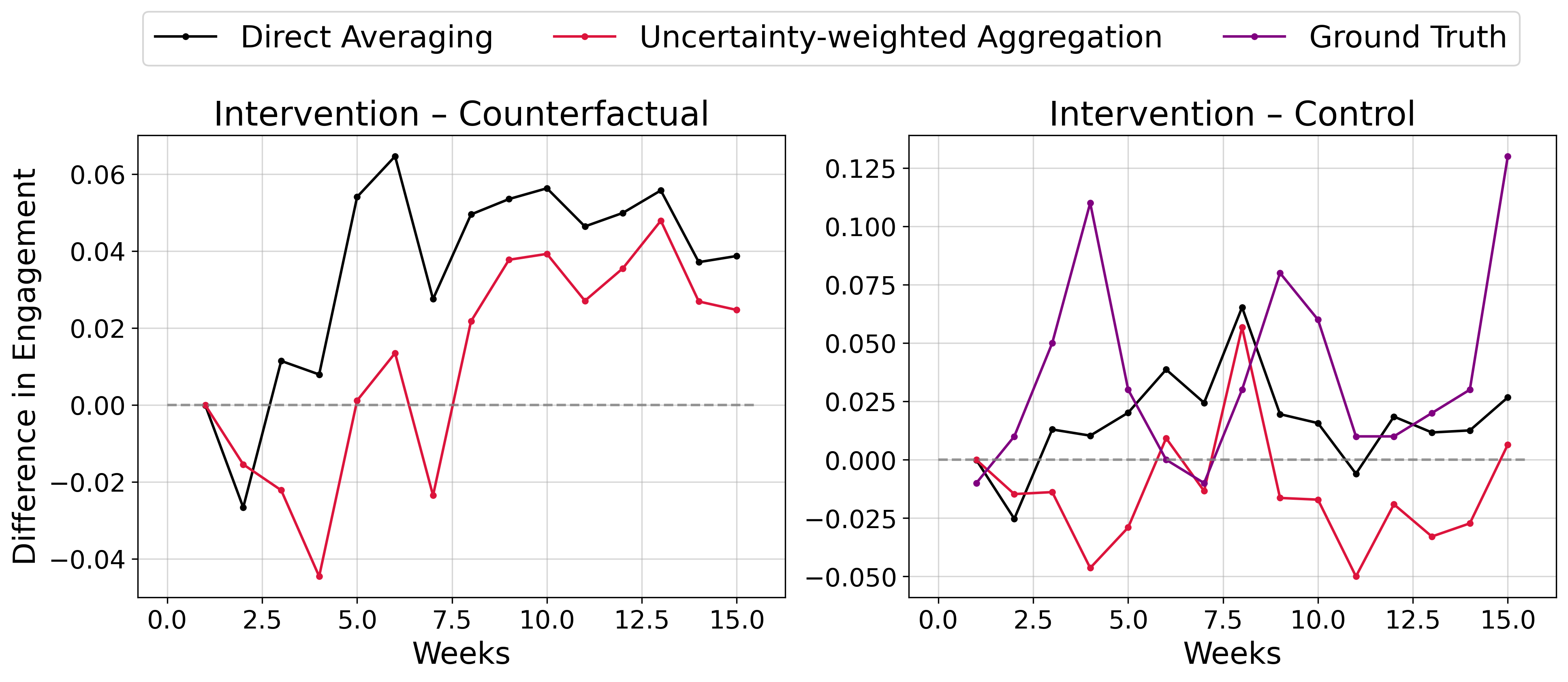}
    \caption{Difference in mean engagement over time between \textit{intervention} and \textit{counterfactual} settings (left), and \textit{intervention} and \textit{control} settings (right).}
    \label{fig:engagement_diff_over_time}
    \Description{}
\end{figure}

Among the three settings, the \textit{intervention} case exhibits the largest prediction variances, indicating greater uncertainty in engagement trajectories. The \textit{counterfactual} engagement closely mirrors the \textit{intervention} case. To quantify predicted effects of live call interventions, we compute differences in mean engagement between the \textit{intervention} setting and the two no-intervention settings (Figure~\ref{fig:engagement_diff_over_time}).

The difference between the \textit{intervention} and \textit{counterfactual} settings (left) is initially negative for both aggregation methods but increases over time. This suggests engagement in the intervention group does not immediately exceed counterfactual expectations, possibly due to model uncertainty in early predictions or a delay in the effect of live calls. However, in later weeks, engagement in the intervention setting surpasses the counterfactual trajectory, suggesting a growing intervention effect.

In contrast, the difference between \textit{intervention} and \textit{control} settings (right) follows more complex trajectories. Ground truth differences (purple) suggest an overall positive effect of live service call interventions, though engagement fluctuates over time. The high variability in the ground truth may reflect real-world fluctuations in engagement patterns, such as seasonal effects, external influences on participation, or heterogeneity in how different mothers respond to live calls. Model predictions do not fully capture this trend. While direct averaging (black) aligns more closely with the true improvement in engagement, uncertainty-weighted aggregation (red) tends to predict smaller differences. This likely arises because uncertainty-weighted aggregation overestimates engagement in the control setting (Figure~\ref{fig:engagement_over_time}), leading to an underestimation of relative intervention benefit. This may be due to the weighting mechanism amplifying predictions from models that are overconfident yet miscalibrated in control settings.

\paragraph{Transitions}

\begin{figure}[h]
    \centering
    \includegraphics[width=\linewidth]{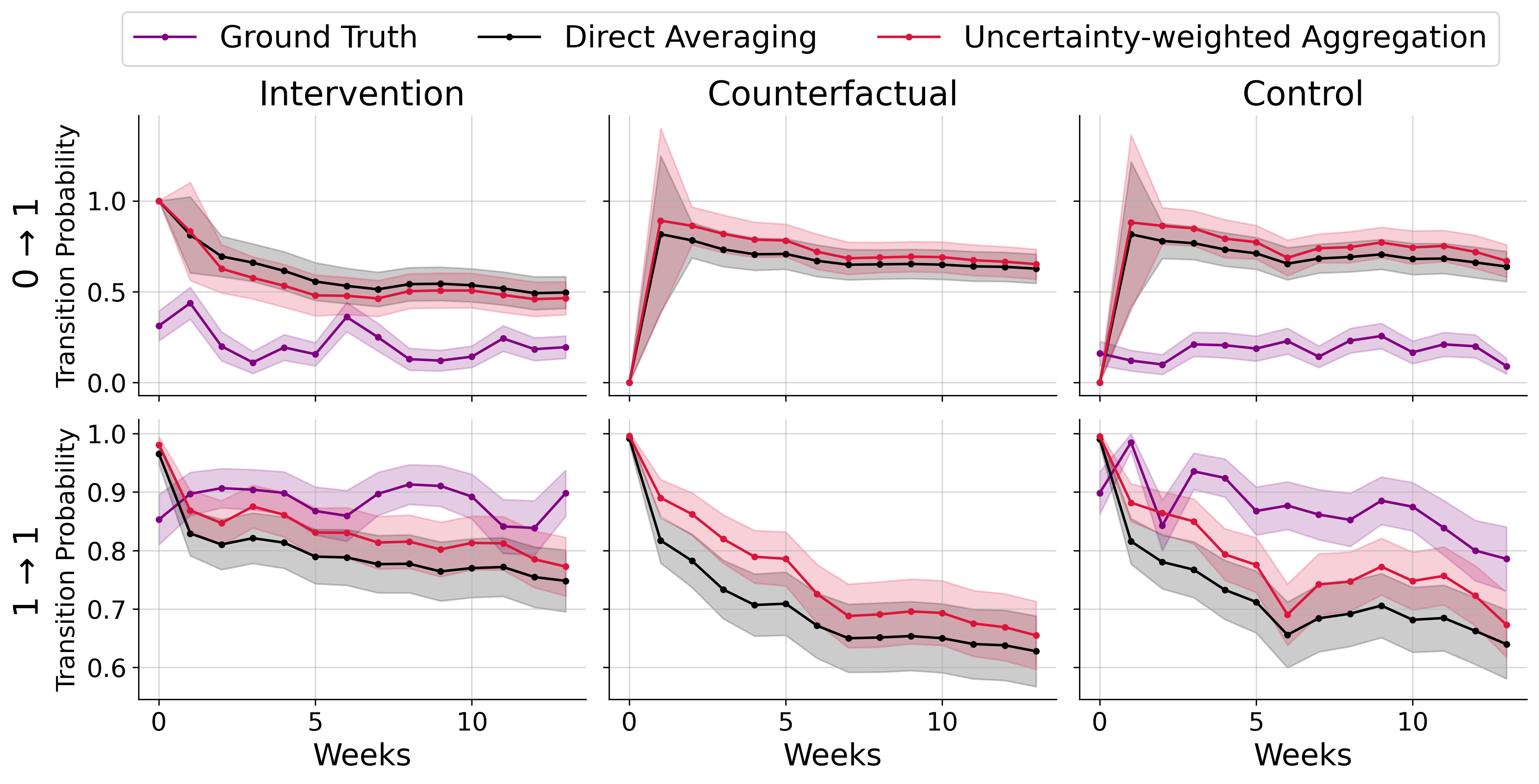}
    \caption{Transition probabilities over time for engagement states across \textit{intervention}, \textit{counterfactual}, and \textit{control} settings. Top Row: Transitions from not engaged (0) to engaged (1). Bottom Row: Probability of remaining engaged (1 → 1).}
    \Description{}
    \label{fig:transition_probs}
\end{figure}

For a more fine-grained analysis of the intervention’s effects on engagement, we examine transition probabilities between `engaged' and `not engaged' states over time (Figure~\ref{fig:transition_probs}) to distinguish between retention (sustained engagement) and re-engagement (recovering previously disengaged users).

The \textit{counterfactual} and \textit{control} settings exhibit similar transition trends, suggesting that, in the absence of direct intervention, predicted engagement behavior follows a comparable trajectory across these settings, as expected. The similarity between \textit{control} and \textit{counterfactual} settings confirms that engagement trends without intervention are largely stable, validating that counterfactual predictions approximate a no-intervention scenario.  

Across settings, aggregation method predictions tend to overestimate both engagement transitions (0 → 1) and retention (1 → 1) relative to ground truth. However, in the \textit{control} setting, models underestimate retention (1 → 1). In the \textit{intervention} setting, the probability of transitioning from not engaged (0) to engaged (1) is lowest long-term among the three settings. While the intervention initially boosts engagement, its predicted effect on re-engagement diminishes over time. This suggests that live calls primarily help sustain engagement rather than recover disengaged users. Conversely, retention probabilities (1 → 1) are highest in the intervention setting, indicating that once engaged, users are more likely to stay engaged with live call interventions in early weeks. Considering both predictions and ground truth trends, service calls as an intervention appear to be more effective at sustaining engagement (1 → 1) than at driving new or re-engagement (0 → 1).


%% file: Sections/5-conclusion.tex
\section{Discussion and Conclusions}

Our findings suggest that LLMs can serve as effective predictive tools for engagement modeling in maternal health programs, particularly when combined through ensemble aggregation methods. However, our analysis also highlights key challenges, particularly regarding overconfidence in LLM predictions, which has been empirically observed in other settings \cite{xiong2024llmuncertainty}.

While Gemini Flash performed well as an individual model in terms of accuracy and F1 score, aggregation methods demonstrated their strength in handling variability and uncertainty. These methods contribute to improved stability and generalization, especially in scenarios where individual models show greater fluctuations or overconfidence. F1 score of aggregated models is never outperformed by any component model, reinforcing benefits of ensemble methods in balancing precision and recall. Model aggregation has the greatest impact on log-likelihood, with direct averaging achieving slightly better log-likelihood values than uncertainty-weighted aggregation. This highlights the robustness of aggregation over any single model’s prediction, improving probability calibration. Ensemble aggregation improves predictive robustness, particularly in settings with data sparsity, by leveraging model diversity. 

Counterfactual predictions provide a valuable tool for intervention analysis, allowing us to simulate and compare engagement trends across settings. Findings indicate that interventions primarily sustain engagement (1 → 1) rather than drive re-engagement (0 → 1). Our results suggest LLMs can be a useful decision-support tool, but they require calibration and aggregation to mitigate overconfidence and ensure reliability. Specifically, direct application of individual LLM predictions may lead to biased intervention planning because of overconfident engagement forecasts—in both cases, but specifically the control setting, we observe an optimistic bias in predicted engagement. In general, a conservative corrective adjustment should be used when interpreting results, and corrective methods should be further explored to mitigate this trend.

Improving direct participant matching between control and intervention groups—ensuring more comparable populations—could help equalize systematic bias across groups, enabling more accurate estimation of difference in engagement (Figure~\ref{fig:engagement_diff_over_time}). When initial LLM-based predictions appear promising, iterative small-scale pilot trials can be used to recalibrate model outputs, yielding more realistic estimates of intervention effects while still maintaining a significantly lower cost than full-scale randomized controlled trials. These `mini pilots' could also help address prediction degradation observed over extended time horizons (Figures~\ref{fig:acc_vs_week} and~\ref{fig:f1_vs_week}). Incorporating real or better-calibrated engagement values into the autoregressive setup—for example, through periodic re-prompting using intermediate outcomes from these trials—may reduce this observed long-term predictive drift. This mirrors adaptive modeling strategies in live trial settings, providing a realistic method for maintaining predictive accuracy over time and better understanding long-term intervention effects. 

\subsection{Future Work} 

Key directions for future work include expanding the counterfactual modeling framework by incorporating sociodemographic matching between participants in control and intervention groups, enabling more explicit and structured comparisons. Further investigations should assess whether models with access to broader contextual intervention descriptions produce more reliable counterfactual predictions than those without, to understand the role of procedural knowledge in behavior forecasting. Another important extension involves developing adaptive weighting schemes for ensembling, where model contributions are dynamically adjusted based on confidence, historical performance, and uncertainty—especially valuable in cases of disagreement among component models. Finally, testing the framework under more realistic conditions—such as simulating program scale-up over time or integrating time-varying sociodemographic features—would provide more detailed insights into how these models and ensembles perform in different real-world deployment scenarios.

Our work demonstrates that LLMs, when properly aggregated, can provide meaningful engagement predictions to guide maternal health interventions. However, challenges related to optimistic bias, long-term prediction degradation, model calibration, and counterfactual prediction reliability remain key areas for future research. By addressing these limitations, LLM-driven approaches could play a significant role in scalable, data-efficient decision-making for social good programs.

\newpage

\subsection*{Consent and Data Usage}

Consent for participating in ARMMAN's mMitra program is received from all beneficiaries. All data collected through the program is owned by ARMMAN and only they are allowed to share data. This dataset will never be used by Google for any commercial purposes. All data used for this project was entirely anonymized before being parsed to any language model; no personally identifiable information is used. Data exchange and use was regulated through clearly defined exchange protocols including anonymization, read-access only to researchers, restricted use of the data for research purposes only, and approval by an ethics review committee registered with the Indian Council of Medical Research.